\documentclass{ifacconf}

\usepackage{graphicx}
\usepackage{natbib} 

\usepackage{amsmath}
\DeclareMathOperator{\dd}{\mathrm{d}}

\DeclareMathOperator{\lt}{lt}
\usepackage{bm}
\usepackage{nicefrac} 

\usepackage{xcolor}

\usepackage{tikz}
\usetikzlibrary{trees,positioning}

\usepackage{url}
\usepackage{subcaption}
\begin{document}
\makeatletter
\g@addto@macro\ps@copyright{%
  \def\@oddfoot{\scriptsize\textcopyright~2026 the authors. Accepted to IFAC for publication under a Creative Commons Licence CC-BY-NC-ND.\hfil}%
  \let\@evenfoot\@oddfoot}
\makeatother
	\begin{frontmatter}
		
		\title{Bayesian Symbolic Regression \\ for Missing Physics} 
		
		\author[First,Second]{Arno Strouwen}
		
		\address[First]{Strouwen Statistics (e-mail: contact@arnostrouwen.com)}
		\address[Second]{Biosystems Department, 
			KULeuven, Leuven, Belgium}
		\begin{abstract} 
			Model-based approaches for (bio)process systems often suffer from incomplete knowledge of the underlying physical, chemical, or biological laws.
			Universal differential equations, which embed neural networks within differential equations, have emerged as powerful tools to learn this missing physics from experimental data.
			However, neural networks are inherently opaque, motivating their post-processing via symbolic regression to obtain interpretable mathematical expressions.
			Genetic algorithm-based symbolic regression is a popular approach for this post-processing step, but provides only point estimates and cannot quantify the confidence we should place in a discovered equation.
			We address this limitation by applying Bayesian symbolic regression, which uses Reversible Jump Markov Chain Monte Carlo to sample from the posterior distribution over symbolic expression trees.
			This approach naturally quantifies uncertainty in the recovered model structure.
			We demonstrate the methodology on a Lotka-Volterra predator-prey system and then show how a well-designed experiment leads to lower uncertainty in a fed-batch bioreactor case study.
		\end{abstract}
		
		
	\end{frontmatter}

	\section{Introduction}
	Model-based approaches are fundamental to the analysis, control, and optimization of (bio)process systems.
	These models encode knowledge of physical, chemical, and biological laws, such as conservation laws, transport phenomena, and reaction kinetics.
	Such knowledge is typically expressed as a system of nonlinear differential equations.
	
	In practice, however, our knowledge of the governing laws is often incomplete.
	These knowledge gaps are termed missing physics and must be inferred from experimental data \citep{harlim}.
	
	Universal Differential Equations (UDEs) have recently emerged as a powerful framework for learning missing model structure \citep{rackauckas1}.
	UDEs embed neural networks within differential equation systems to represent terms whose underlying form is unknown \citep{dandekar}.
	
	The opaque nature of neural networks is often undesirable in scientific computing, where interpretability and physical insight are valued.
	Consequently, UDE-based approaches are frequently combined with interpretable machine learning techniques, such as sparse regression \citep{kaiser} or symbolic regression \citep{koza}.
	These techniques distill the neural network into human-understandable mathematical expressions.
	
	Symbolic regression searches the space of mathematical expressions to find models that balance accuracy and simplicity.
	Mathematical expressions are typically represented as tree data structures, where branch nodes correspond to operators and leaf nodes to variables or constants.
	
	UDE-based methods are rapidly gaining traction across scientific domains, including physics \citep{keithLearningOrbitalDynamics2021}, chemistry \citep{santanaEfficientHybridModeling2023}, and biology \citep{philippsNonNegativeUniversalDifferential2024, rojas-camposLearningCOVID19Regional2023}.
	
	When symbolic regression is employed in these applications, genetic algorithms are the dominant approach for exploring the space of candidate expression trees \citep{cranmer}.
	While genetic algorithms often yield good point estimates for the missing physics, they provide no principled way to quantify uncertainty in the discovered model structure.
	
	Recently, symbolic regression has been formulated within a Bayesian framework by \citet{jin}.
	This approach replaces genetic algorithms with Reversible Jump Markov Chain Monte Carlo (RJMCMC) to explore the space of possible model structures.
	RJMCMC extends standard MCMC by allowing moves between parameter spaces of different dimensions.
	This capability is essential for symbolic regression because different expression trees contain different numbers of constant parameters.
	
	The likelihood of a candidate tree is evaluated by comparing its predictions to the outputs of the trained neural network, while a prior distribution over trees penalizes depth to encourage parsimonious expressions.
	The RJMCMC sampler uses the likelihood and the prior to generate samples from the posterior distribution over symbolic trees, naturally quantifying structural uncertainty.
	The merits of this choice versus computing the likelihood directly from experimental data are discussed in Section~\ref{sec:discussion}.
	In this paper, we apply Bayesian symbolic regression to the missing physics problem for the first time, enabling principled uncertainty quantification over recovered model structures.
	
	We use the Lotka-Volterra predator-prey model to illustrate the concept of recovering missing physics with Bayesian symbolic regression. A fed-batch bioreactor case study is used to show how a well-designed experiment leads to lower uncertainty on the recovered missing physics.
	
	\section{Motivating example: Lotka-Volterra}
	We illustrate the missing physics problem using the Lotka-Volterra predator-prey model, a classical system in mathematical ecology, and a popular tool to illustrate scientific machine learning concepts \citep{rackauckas1}.
	The model describes the dynamics of two interacting populations: prey $x_1$ and predators $x_2$.
	The full system is given by:
	\begin{equation}\label{eq:lv}
		\begin{aligned}
			\frac{dx_1}{dt} &= 1.3 x_1 - 0.9 x_1x_2,\\
			\frac{dx_2}{dt} &= -1.8 x_2 + 0.8 x_1x_2.\\
		\end{aligned}
	\end{equation}
	Suppose we know the linear dynamics but lack knowledge of the interaction terms.
	The incomplete model takes the form:
	\begin{equation}\label{eq:lv-missing}
		\begin{aligned}
			\frac{dx_1}{dt} &= 1.3 x_1 + \phi_1(x_1,x_2),\\
			\frac{dx_2}{dt} &= -1.8 x_2 + \phi_2(x_1,x_2),\\
		\end{aligned}
	\end{equation}
	where $\phi_1(x_1,x_2)$ and $\phi_2(x_1,x_2)$ represent the unknown interaction terms that must be inferred from data.
	
\begin{figure}[htbp]
		\centering
		\includegraphics[width=\columnwidth]{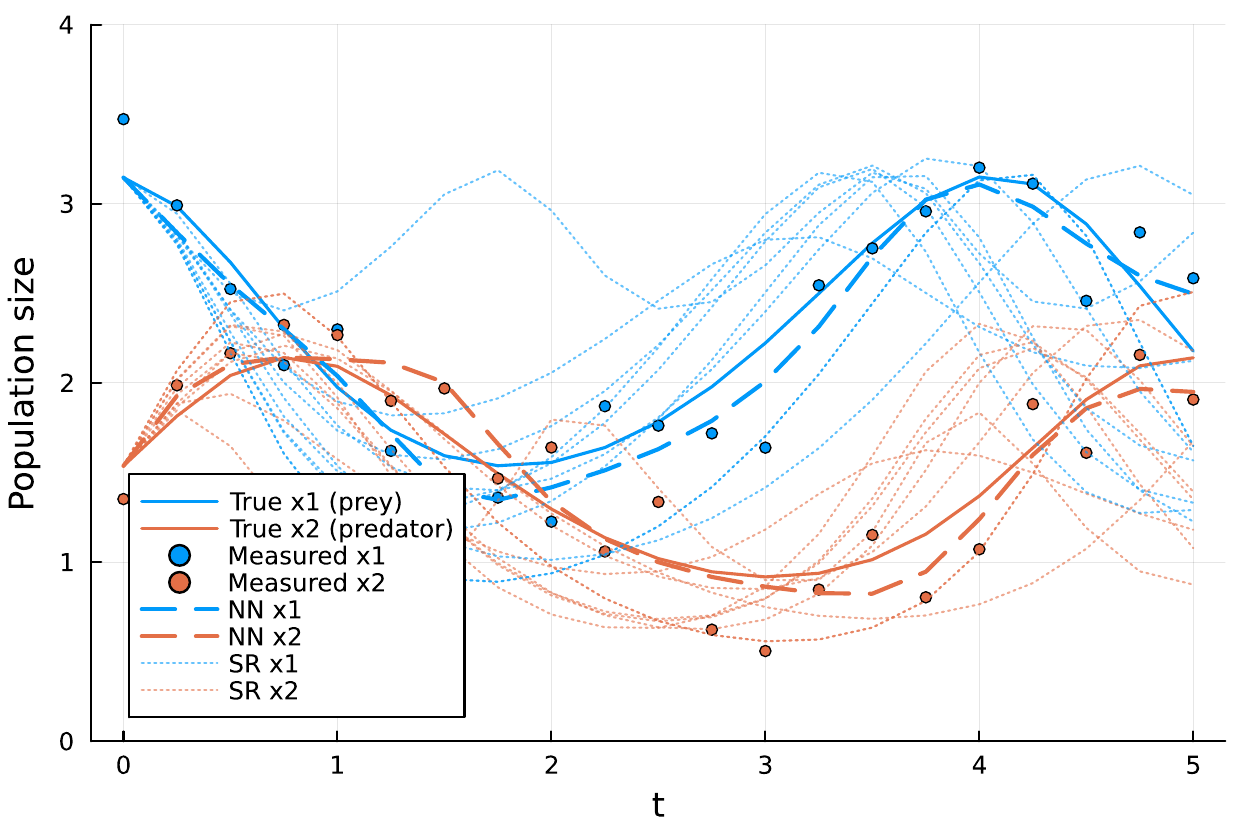}
		\caption{Lotka-Volterra system. Solid lines: true state values. Thick dots: noisy measurements of states. Dashed lines: predicted states after replacing missing physics with a neural network.
			Dotted lines: predicted states after replacing missing physics with 10 draws from Bayesian symbolic regression.}
		\label{fig:lv}
	\end{figure}

	Figure~\ref{fig:lv} shows the state trajectories of the Lotka-Volterra system, and noisy measurements thereof,
	and illustrates how a neural network and Bayesian symbolic regression can be used to fill in the missing physics terms.
	In particular, 10 plausible symbolic equations have been generated.
	These equations are shown in Table~\ref{tab:lv-results}.
	
	\begin{table}[htbp]
		\centering
		\caption{Posterior samples from Bayesian symbolic regression for the Lotka-Volterra missing physics. Here $\lt(x,a,b)=ax+b$.}
		\label{tab:lv-results}
		\resizebox{\columnwidth}{!}{%
		\begin{tabular}{lll}
			\hline
			Term & Expression & Chains \\
			\hline
			$\phi_1$ & $-(x_1 x_2)$ & 5 \\
			(true: $-0.9x_1x_2$) & $(x_1^2 x_2) / -(x_1)$ & 1 \\
			& $-(x_1) / (x_1 / (x_1 x_2))$ & 1 \\
			& $x_1 + ((x_2 x_1 + x_1) / (-(x_2) / x_2))$ & 1 \\
			& $-\lt(x_2 x_1, 0.95, -0.18)$ & 1 \\
			& $\lt(x_2, 0.35, -0.46) + x_1^2 / (((x_1/x_2)/-(x_1)) \cdot x_1)$ & 1 \\
			\hline
			$\phi_2$ & $x_1 x_2$ & 2 \\
			(true: $0.8x_1x_2$) & $x_2 \cdot \lt(x_1, a, b)$ & 3 \\
			& $\lt(x_1 x_2, 0.89, -0.18)$ & 1 \\
			& $x_2 \cdot (\exp(x_1) / ((x_1 + x_1) + x_2))$ & 1 \\
			& $(x_2 / x_1) / -(x_1) + x_2 x_1$ & 1 \\
			& $(-(x_2 + x_2) / \exp(x_1) + x_2) \cdot x_1$ & 1 \\
			& $-\lt(\exp(x_1), -0.20, 1.48) \cdot x_2$ & 1 \\
			\hline
		\end{tabular}%
		}
	\end{table}
	The interaction term between predator and prey occurs often in these equations.
	For $\phi_1$, 8 out of 10 chains recovered an expression algebraically equivalent to $-x_1 x_2$.
	The same equation can appear in many different forms, e.g., $(x_1^2 x_2)/-(x_1)$ simplifies to $-x_1 x_2$.
	For $\phi_2$, the majority of chains contain the $x_1 x_2$ interaction, though some express it through nonlinear functions such as $\exp(x_1) \cdot x_2$.
	In some equations, the coefficients $0.9$ and $0.8$ are also approximately recovered,
	while for $\phi_1$ the most common outcome is a parsimonious equation without constants.
	
	In the following sections, we present the methodology used to recover these terms.
	
	\section{Universal differential equations}
	We consider dynamical systems of the following form:
	\begin{equation}\label{eq:system}
		\begin{aligned}
			\frac{\dd \bm x}{\dd t} &= \bm f(t,\bm x,\bm \phi(\bm g(\bm x))), \qquad \text{with } \bm x(t=0) = \bm x_0;\\
			\bm y_k &= \bm h(\bm x(t_k)) + \bm \epsilon_k,
		\end{aligned}
	\end{equation}
	where $t$ denotes time and $\bm x(t)$ is the state vector.
	The states evolve according to the ordinary differential equation system $\bm f$ with initial conditions $\bm x_0$.
	Measurements $\bm y_k$ are collected at discrete time points $t_k$ and are related to the states through the observation function $\bm h$.
	This measurement function is useful, for example, when only a subset of states is observed.
	The measurements are corrupted by independent Gaussian noise $\bm \epsilon_k$, with each $\bm \epsilon_k$ identically and independently distributed as multivariate normal with zero mean and covariance matrix $R$.
	
	The system $\bm f$ depends on the unknown function $\bm \phi$, which represents the missing physics.
	The input to $\bm \phi$ is the output of a known preprocessing function $\bm g(\bm x)$.
	This function $\bm g$ is useful when the missing physics depends only on a subset of the states or on known transformations thereof.
	We refer to $\bm g(\bm x)$ as the features.
	
	In the UDE framework, the unknown function $\bm \phi$ is replaced by a neural network:
	\begin{equation}\label{eq:UDE}
		\frac{\dd \bm x}{\dd t} = \bm f(t,\bm x,\text{NN}(\bm g(\bm x), \bm \theta)).
	\end{equation}
	The neural network $\text{NN}(\bm g(\bm x), \bm \theta)$ takes the features $\bm g(\bm x)$ as input and depends on parameters $\bm \theta$ that must be learned from experimental data.
	Training of $\bm \theta$ proceeds by minimizing the mean squared error of the observations.

	\section{Bayesian symbolic regression}
	We want to find mathematical expressions that match the output of the neural network and quantify how certain we are a specific expression is correct.
	
\begin{figure}[htbp]
		\centering
		\begin{tikzpicture}[
			level distance=1.2cm,
			sibling distance=2.5cm,
			every node/.style={draw, circle, minimum size=8mm, inner sep=1pt},
			edge from parent/.style={draw, -},
			level 1/.style={sibling distance=3cm},
			level 2/.style={sibling distance=1.2cm},
			feature/.style={draw, rectangle, rounded corners, fill=gray!20},
			const/.style={draw, rectangle, fill=white}
			]
			\node {$+$}
			child {node {$\lt$}
				child {node[feature] {$x_1$}}
				child {node[const] {$2$}}
				child {node[const] {$1$}}
			}
			child {node {$\sin$}
				child {node[feature] {$x_2$}}
			};
		\end{tikzpicture}
		\caption{Expression tree representing $(2x_1+1) + \sin(x_2)$. Circular nodes contain operators ($+$, $\lt$, $\sin$). Rectangular nodes are terminals: rounded rectangles denote features ($x_1$, $x_2$), and plain rectangles denote constants. The $\lt$ node implements the linear transformation $\lt(x_1,2,1) = 2x_1 + 1$.}
		\label{fig:tree}
	\end{figure}
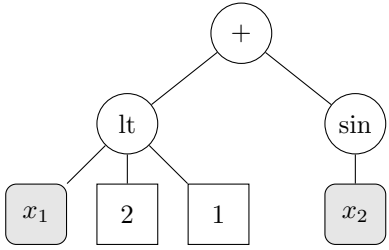

	To formalize this inference problem, we represent mathematical expressions as tree data structures, as illustrated in Figure~\ref{fig:tree}.
	The (non-terminal) branch nodes of the tree contain mathematical operators,
	while the (terminal) leaf nodes contain either features or constants.
	Each branch node has one or two children, depending on its arity, i.e., whether it is a unary or binary mathematical operator.
	All leaf nodes contain features except a special linear transformation operator $\lt(x,a,b)=ax+b$, here the second and third children, $a$ and $b$ are constants, and the first child $x$ is either an operator or a feature.
	This $\lt$ operator is the only way for constants, such as 0.9 and 0.8 in the Lotka-Volterra example, to enter the mathematical expression.
	
	To perform Bayesian inference, we must specify a prior distribution over trees and a likelihood function.
	\subsection{Prior}
	The prior over trees is constructed from several component distributions.
	The \textit{terminal prior} specifies the probability that a node at a given depth is a branch node.
	This probability decreases with depth to discourage overly complex trees.
	The \textit{feature prior} assigns probabilities to different features at leaf nodes.
	The \textit{operator prior} assigns probabilities to different operators at branch nodes.
	Finally, \textit{priors for $a$ and $b$} specify distributions over the constants in $\lt$ nodes.
	
	Using these distributions, the prior over trees is defined recursively.
	Starting at the root (depth zero), the terminal prior determines whether the node is a leaf or branch node.
	If the node is a leaf, a feature is sampled from the feature prior.
	If the node is a branch, an operator is sampled from the operator prior.
	If this operator is $\lt$, then constants $a$ and $b$ are also sampled from their respective priors.
	Child nodes for this operator are then generated recursively with incremented depth.
	
	Sampling a tree from the prior thus involves both discrete choices (whether nodes are leaves or branches, selection of features and operators) and continuous choices (the constants $a$ and $b$ in $\lt$ nodes).
	We denote the collection of discrete choices by $T$ and the continuous choices by $A$ and $B$, which contain the $a$ and $b$ constants when traversing the tree in depth-first order.
	Note that the dimensions of $A$ and $B$ vary across trees, as different trees contain different numbers of $\lt$ nodes.
	
	The tree will not fit the neural network output exactly.
	We assume the discrepancy follows an additive i.i.d. normal distribution with variance $\sigma^2$.
	A \textit{variance prior} on $\sigma$ completes the Bayesian specification.
	
	We use the notation $p(T, A, B, \sigma)$ to denote the prior density for a given tree.
	\subsection{Likelihood}
	Let $T(\bm g(\bm x_k), A, B)$ denote the output of a tree evaluated at the features $\bm g(\bm x_k)$ with constants $A$ and $B$.
	Since each tree produces a scalar output, a neural network with $m$ outputs requires $m$ separate trees $T_1, \ldots, T_m$, each with its own structure and constants $(T_j, A_j, B_j)$.
	We denote the $j$-th output of the neural network as $\text{NN}_j(\bm g(\bm x_k), \bm \theta)$; each tree $T_j$ should approximate the corresponding output.
	Since the trees for different neural network outputs are independent, we run a separate Markov chain for each output $j$.
	Because the inference for each tree is handled separately, in the following, we drop the subscript $j$ for clarity.
	
	The log-likelihood at measurement $k$ is:
	\begin{equation}\label{eq:likelihood}
		\begin{aligned}
			&\log p(\text{NN}(\bm g(\bm x_k), \bm \theta) | T, A, B, \sigma) =\\
			&\quad -\frac{1}{2}\log(2\pi\sigma^2) -\frac{\left( T(\bm g(\bm x_k), A, B) - \text{NN}(\bm g(\bm x_k), \bm \theta) \right)^2}{2\sigma^2}.
		\end{aligned}
	\end{equation}
	Assuming independence across time points, the total log-likelihood is the sum over all $k$.
	For notational simplicity, we write this as $p(\text{NN} | T, A, B, \sigma)$, with analogous notation for the posterior.
	\subsection{Markov chain}
	We use hybrid Gibbs sampling of \citet{tierney} with two transition kernels.
	The first kernel updates only the continuous parameters $\sigma$, $A$, and $B$, holding the tree structure $T$ fixed.
	Since these parameters are continuous, we can use efficient gradient-based samplers such as NUTS \citep{hoffman}.
	
	The second kernel updates the discrete tree structure $T$ using Reversible Jump Markov Chain Monte Carlo (RJMCMC) by \citet{green}.
	A high-level explanation of this algorithm is that it proposes random modifications to the tree structure.
	Modifications that increase the posterior probability are more likely to be accepted than modifications that decrease posterior probability. The exact acceptance rate of modifications is calibrated such that the detailed balance is maintained.
	This ensures the stationary distribution of the Markov chain is the posterior distribution.
	
	Changing the tree structure cannot be done independently of the continuous parameters, since $\lt$ nodes may appear or disappear.
	This change in dimensionality of $A$ and $B$ requires careful treatment in RJMCMC.

	Let $K(\sigma, T^*, A^*, B^* | \sigma, T, A, B)$ denote the transition kernel, which defines a probability distribution over the next state $(\sigma, T^*, A^*, B^*)$ given the current state $(\sigma, T, A, B)$.
	Note that $\sigma$ appears on both sides without modification, as this kernel only updates the tree structure and its constants; $\sigma$ is updated separately.
	
	We only consider modifications where $\lt$ nodes are either added or deleted, but not both simultaneously.
	When $\lt$ nodes are added, the new constants $a$ and $b$ are sampled from their respective priors.
	We denote these newly sampled parameters as $u_A$ and $u_B$.
	
	To maintain the detailed balance, we require that:
	\begin{equation}\label{eq:balance}
		\begin{aligned}
			&p(T, A, B, \sigma|\text{NN})\\
			&\quad \times K(\sigma, T^*, A^*, B^* | \sigma, T, A, B)dA^*dB^*=\\
			&p(T^*, A^*, B^*, \sigma|\text{NN})\\
			&\quad \times K(\sigma, T, A, B | \sigma, T^*, A^*, B^*)dAdu_AdBdu_B,
		\end{aligned}
	\end{equation}
	i.e., that the flow from $T, A, B$ to $T^*, A^*, B^*$  matches the reverse flow.
	If a bijection $\psi$ can be established between
	$(A, u_A, B, u_B)$ and $(A^*, B^*)$,
	then we get:
	\begin{equation}\label{eq:substitute}
		dA^*dB^*=\\
		\left | \det \frac{\partial \psi}{\partial A \partial u_A \partial B  \partial u_B } \right |
		dAdu_AdBdu_B
	\end{equation}
	Such a bijection is straightforward to construct: $A^*$ is formed by inserting $u_A$ at positions corresponding to new $\lt$ nodes in depth-first order, and similarly for $B^*$.
	Since this is a permutation, the absolute value of the determinant is $1$, and thus:
	\begin{equation}\label{eq:balance_simplified}
		\begin{aligned}
			&K(\sigma, T^*, A^*, B^* | \sigma, T, A, B) = \\
			&K(\sigma, T, A, B | \sigma, T^*, A^*, B^*)\\
			&\quad \times\frac{p(T^*, A^*, B^*, \sigma|\text{NN})}{p(T, A, B, \sigma|\text{NN})}
		\end{aligned}
	\end{equation}
	
	The same argument applies when $\lt$ nodes are deleted: the removed parameters $u_A$ and $u_B$ can be extracted from $A$ and $B$ via the inverse permutation, and (\ref{eq:balance_simplified}) still holds.
	
	We split the transition kernel into two factors.
	The proposal distribution $q(T^*, A^*, B^* | T, A, B)$ defines the probability of proposing tree $(T^*, A^*, B^*)$ given current tree $(T, A, B)$.
	The acceptance of the proposal is determined by acceptance probability $\alpha(\sigma, T^*, A^*, B^* |\sigma, T, A, B)$.
	If the proposal is rejected, the tree remains unchanged for this iteration.
	
	By setting:
	\begin{equation}
		\begin{aligned}
			&\log \alpha(\sigma, T^*, A^*, B^* | \sigma, T, A, B) = \\
			\min [ 0,\; &\log p(T^*, A^*, B^*, \sigma|\text{NN})\\
			-&  \log p(T, A, B, \sigma|\text{NN})\\
			+& \log q(T, A, B | T^*, A^*, B^*)\\
			-&  \log q(T^*, A^*, B^* | T, A, B)],
		\end{aligned}
	\end{equation}
	equation (\ref{eq:balance_simplified}) is always satisfied.
	In practice, we sample $u \sim \text{Uniform}(0,1)$ and accept the proposal if $\log u < \log \alpha$.
	Note that this derivation mirrors standard random walk Metropolis-Hastings, but with special care for the transdimensional nature of the problem: the bijection $\psi$ accounts for parameters appearing or disappearing when $\lt$ nodes are added or removed.
	
	The proposal distribution $q$ follows \citet{jin} and consists of several move types:
	
	With probability $p_{rf}$, the tree \textbf{reassigns a feature}: a randomly selected leaf node is replaced with a feature sampled from the feature prior.
	The reverse move reassigns the old feature to the same node.
	
	With probability $p_{ro}$, the tree \textbf{reassigns an operator}: a randomly selected branch node is replaced with an operator of the same arity, sampled from the operator prior (conditioned on arity).
	If no branch node exists, the tree is unchanged.
	Note that $\lt$ is the only ternary operator, and thus will remain unchanged,
	and this proposal does not change the $a$ and $b$ children of the $\lt$ node.
	The reverse move reassigns the old operator to the same node,
	(or does nothing if no branch node exists).
	
	With probability $p_p$, the tree is \textbf{pruned}: a randomly selected branch node and its children are replaced with a feature sampled from the feature prior.
	If no branch nodes exist, the tree is unchanged.
	The reverse of pruning is growing.
	
	With probability $p_g$, the tree is \textbf{grown}: a randomly selected leaf node is replaced by a subtree sampled from the prior, conditioned on that selected node's depth.
	To ensure grow is the reverse of prune, the root of the new subtree must be a branch node; if a leaf is sampled as the root, it is resampled until a branch node is obtained.
	
	With probability $p_i$, an operator is \textbf{inserted} between a randomly selected node and its parent.
	That inserted operator is sampled from the operator prior.
	If the inserted operator is unary, the insertion is straightforward.
	If the inserted operator is binary, then another subtree has to be grown,
	that is randomly assigned to either the left or right child of the inserted operator.
	If the inserted operator is $\lt$, then new $a$ and $b$ parameters have to be sampled.
	The reverse of inserting an operator is deleting an operator.
	
	With probability $p_d$, a randomly selected branch node of the tree is \textbf{deleted}.
	If the selected operator node is unary, its parent and child are connected.
	If the selected operator node is binary, its parent and a randomly selected child are connected.
	If the selected operator node is $\lt$, its parent and first child are connected,
	while the second and third children ($a$ and $b$) are discarded.
	
	Of course, $p_{rf} + p_{ro} + p_p + p_g + p_i + p_d$ must equal 1. 
	\section{Computational details}
	The training of the universal differential equations follows \citet{strouwen}.
	The symbolic trees for Bayesian symbolic regression are implemented using DynamicExpressions.jl.
	This package forms the backbone of representing trees in the genetic algorithm based SymbolicRegression.jl \citep{cranmerInterpretableMachineLearning2023} and is repurposed here for Bayesian inference.
	
	The negation and exponential functions are used as unary operators.
	Addition, multiplication, and division are used as binary operators.
	Uniform distributions are used for the operator and feature priors.
	Normal distributions, with standard deviation 2 and means 1 and 0 are used for the priors of $a$ and $b$ parameters respectively.
	A Bernoulli distribution is used as the terminal prior,
	where the probability of being a branch node at depth $d$ equals $0.9(1 + d)^{-0.7}$,
	with the root node having depth zero.
	The noise prior is an Inverse-Gamma distribution with shape parameter 1 and scale parameter 0.001.
	
	To draw from the posterior we run 100 separate Reversible Jump Markov Chains,
	each for 100,000 iterations, and keep the final one.
	Continuous parameters are initialized with 100 NUTS \citep{hoffman} steps.
	During RJMCMC, $\sigma^2$ is updated every iteration via its conjugate Gibbs update, while $\lt$ node constants are refined every 10,000 iterations and after the final iteration using random walk Metropolis-Hastings.
	On a single core of an AMD Ryzen 9 5900X, each chain completes in approximately 2 seconds for the Lotka-Volterra example and 5 seconds for the bioreactor example.
	All chains are independent and thus trivially parallelizable.

	The source code accompanying this paper can be found at \url{https://github.com/arno-papers/IFACWC2026}.
	
	This repository also contains additional examples, such as missing physics of a SIR model for COVID-19 modeling.
	\section{Bioreactor Case Study}
	We illustrate how to use Bayesian symbolic regression to quantify model structure uncertainty using a well-mixed fed-batch bioreactor.
	This reactor has a long history in the experimental design literature \citep{versyck}.
	The reactor is represented by three states:
	the substrate concentration, $C_s$,	the biomass concentration, $C_x$, and the volume of the reactor, $V$.
	The evolution of these states is governed by the system of differential equations:
		\begin{equation}\label{eq:bioreactor}
		\begin{aligned}
			\frac{dC_s}{dt} &= -\left(\frac{\mu(C_s)}{0.777}\right) C_x + \frac{Q_{in}(t)}{V}(50 - C_s),\\
			\frac{dC_x}{dt} &= \mu(C_s) C_x - \frac{Q_{in}(t)}{V}C_x,\\
			\frac{dV}{dt} &= Q_{in}(t).
		\end{aligned}
	\end{equation}
	In these equations, the specific growth rate, $\mu$, is the missing physics.
	This function has a single input, $C_s$.
	The true function that must be recovered is the Monod equation:
	\begin{equation}
		\mu(C_s) = \frac{\mu_{max}C_s}{K_s + C_s}.
	\end{equation}
	The differential equations are also dependent on a controllable input $Q_{in}(t)$.
	
\begin{figure}[htbp]
		\centering
		\includegraphics[width=\columnwidth]{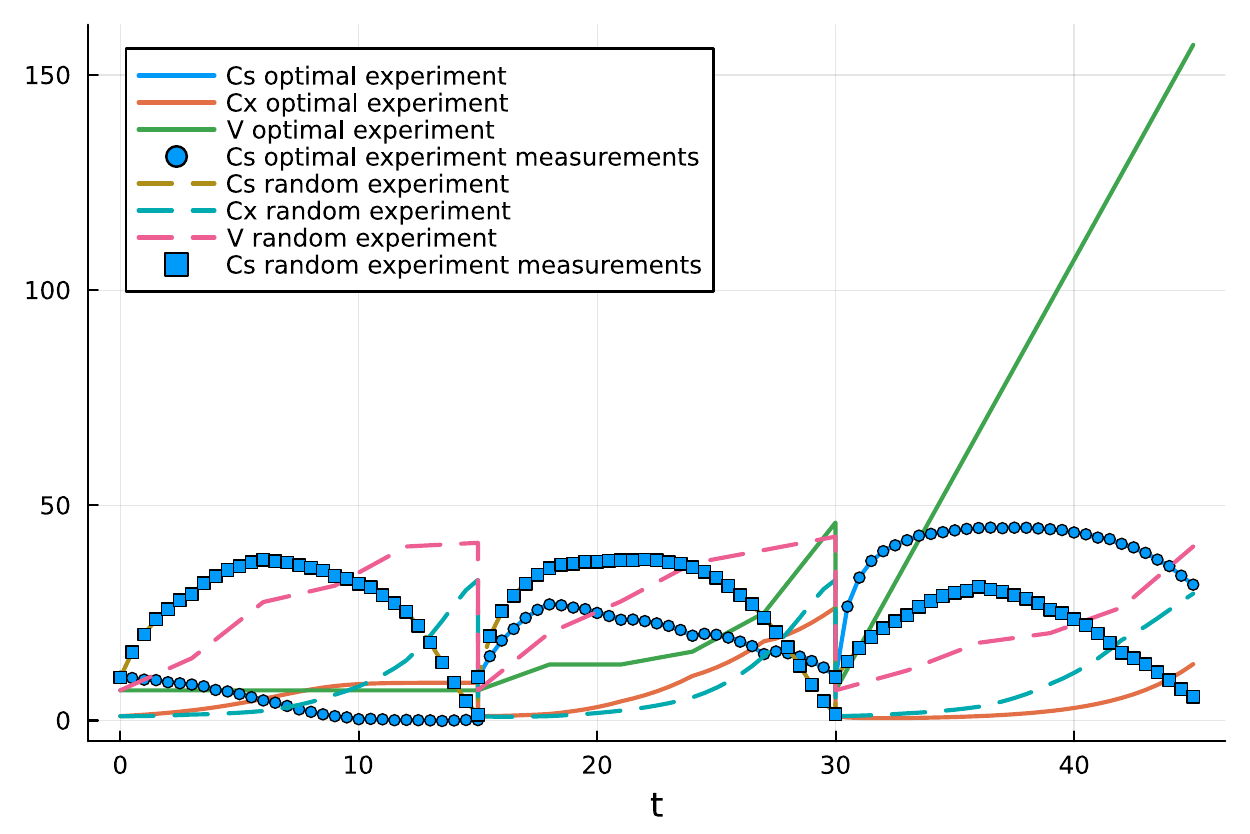}
		\caption{Bioreactor system. States and measurements.}
		\label{fig:biostates}
	\end{figure}

\begin{figure}[htbp]
		\centering
		\begin{subfigure}[b]{\columnwidth}
			\centering
			\includegraphics[width=\columnwidth]{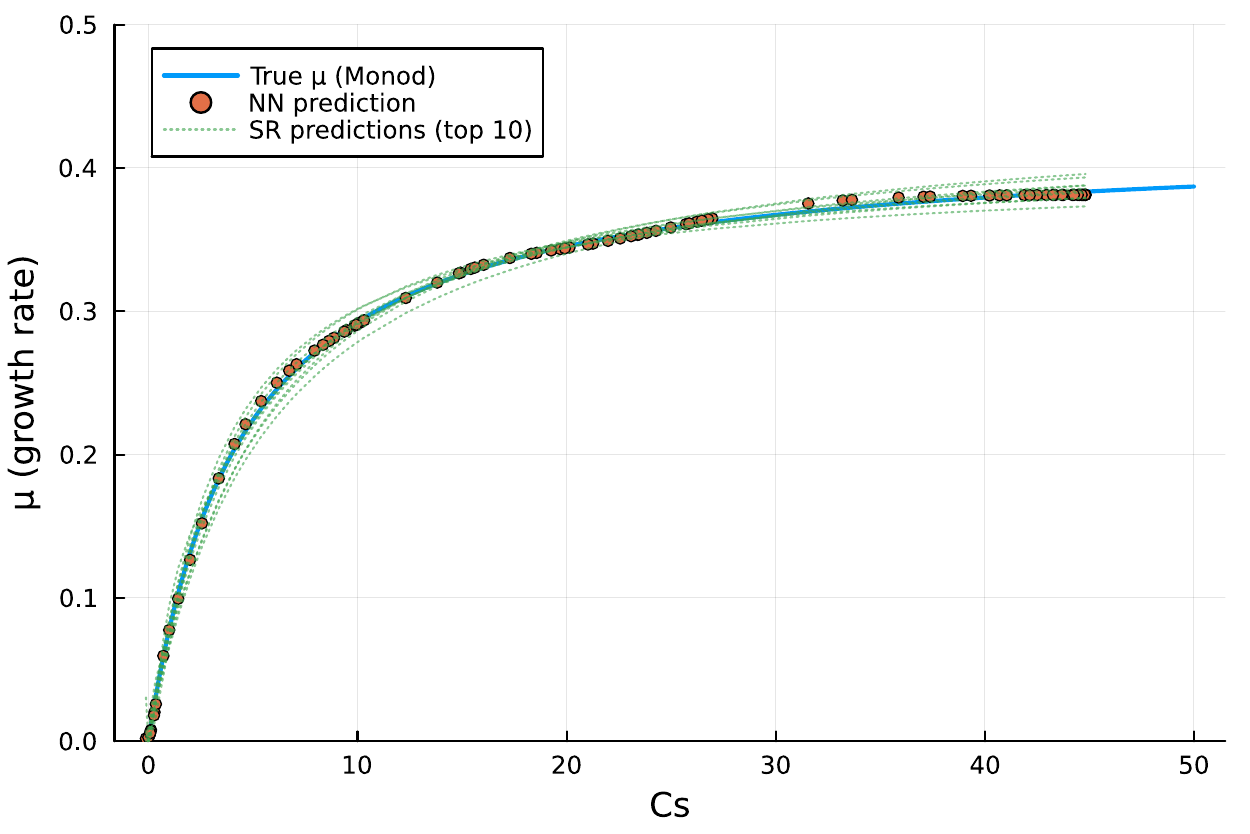}
			\caption{Optimal experimental design: the equations lay in a tight region around the true missing physics, particularly at low substrate concentrations.}
			\label{fig:bioreactor_optimal}
		\end{subfigure}
		
		\begin{subfigure}[b]{\columnwidth}
			\centering
			\includegraphics[width=\columnwidth]{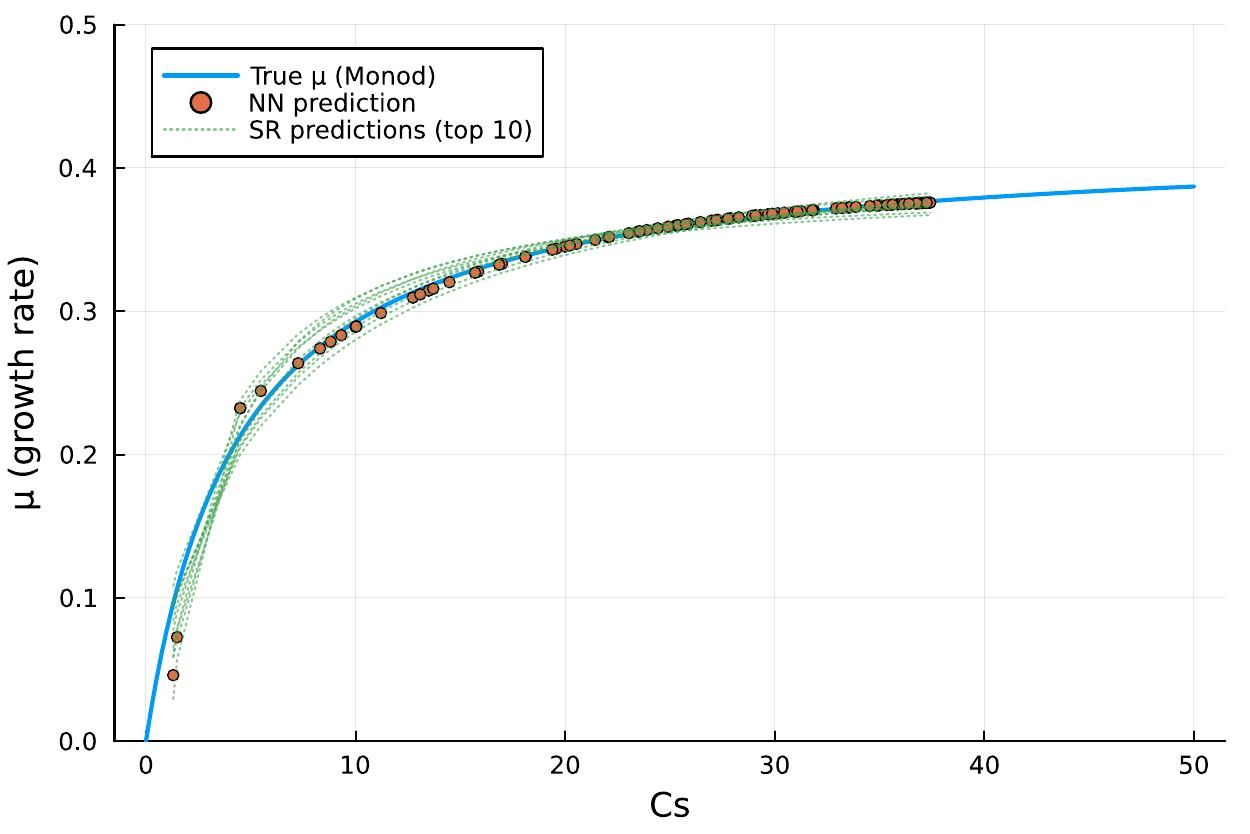}
			\caption{Random experimental design: the equations exhibit a wider spread, especially at low and medium substrate concentrations where fewer measurements are collected.}
			\label{fig:bioreactor_random}
		\end{subfigure}
		\caption{Fed-batch bioreactor system. Blue solid line: missing physics, the Monod equation. Thick orange dots: predictions of the missing physics by the neural network. Green dashed lines: predicted missing physics by the top 10 trees generated by Bayesian symbolic regression.}
		\label{fig:bioreactor}
	\end{figure}

	Recently, \citet{strouwen} constructed an optimal experimental design, in which
	$Q_{in}(t)$ was optimized to recover $\mu$ as efficiently as possible, from noisy $C_s$ measurements.
	The optimal experiment can recover the Monod equation via genetic algorithm based symbolic regression,
	while for the random experiment, where the control takes on random values, the correct equation is not recovered.
	Figure~\ref{fig:biostates} depicts the system both for optimal and random controls.
	We now apply Bayesian symbolic regression to this data.
	The raw expression trees are complex, but algebraic simplification reveals a clear pattern.
	As shown in Table~\ref{tab:bioreactor_equations}, four of five optimal equations reduce to the Monod form $C_s / (\alpha C_s + \beta)$, with estimated parameters closely matching the true values.
	None of the random equations recover the Monod form: three simplify to $C_s^2 / (\alpha C_s^2 + \beta C_s + \gamma)$, exhibiting quadratic rather than the correct linear behavior at low $C_s$, while the remaining two contain non-polynomial terms.
	Figure~\ref{fig:bioreactor} confirms that this discrepancy is most pronounced at low substrate concentrations, where the random experiment collects fewer measurements.
	\begin{table}[htbp]
		\centering
		\caption{Algebraically simplified equations for the bioreactor with estimated Monod parameters. The optimal experiment recovers the Monod form, while the random experiment does not.}
		\label{tab:bioreactor_equations}
		\begin{tabular}{llll}
			\hline
			& Simplified form & $\hat{\mu}_{\max}$ & $\hat{K}_s$ \\
			\hline
			Opt.\ 1 & $C_s / (2.33\, C_s + 11.09)$ & 0.43 & 4.77 \\
			Opt.\ 2 & non-polynomial ($\exp(-C_s)$ terms) & --- & --- \\
			Opt.\ 3 & $C_s / (2.43\, C_s + 9.41)$  & 0.41 & 3.88 \\
			Opt.\ 4 & $C_s / (2.38\, C_s + 10.83)$ & 0.42 & 4.55 \\
			Opt.\ 5 & $C_s / (2.25\, C_s + 12.44)$ & 0.44 & 5.53 \\
			\hline
			True & $\mu_{\max}\, C_s / (K_s + C_s)$ & 0.421 & 4.39 \\
			\hline
			Rand.\ 1 & $C_s^2 / (2.46\, C_s^2 + 7.50\, C_s + 7.96)$ & --- & --- \\
			Rand.\ 2 & non-polynomial ($\exp(1/C_s)$ terms) & --- & --- \\
			Rand.\ 3 & $C_s^2 / (2.57\, C_s^2 + 4.92\, C_s + 18.10)$ & --- & --- \\
			Rand.\ 4 & $C_s^2 / (2.34\, C_s^2 + 11.76\, C_s - 1.47)$ & --- & --- \\
			Rand.\ 5 & non-polynomial ($1/C_s$ terms) & --- & --- \\
			\hline
		\end{tabular}
	\end{table}

	\section{Discussion}\label{sec:discussion}
	At the start of this research project, we aimed to eliminate the neural network as an intermediate step and directly compute the likelihood of the trees based on the measured observations using (\ref{eq:system}), rather than the neural network output as in (\ref{eq:likelihood}).
	However, this proved computationally infeasible: evaluating symbolic trees within a differential equation solver introduces significant overhead that DynamicExpressions.jl is currently not optimized for. Another issue is that RJMCMC suffers from the curse of dimensionality, similar to random walk Metropolis-Hastings. We hope that advances in MCMC methods for discrete parameters may make this direct approach feasible in the future \citep{gagnon, power}.

	Removing the neural network intermediate would bring additional benefits.
	Currently, the trees are trained separately for each neural network output,
 	with no joint uncertainty quantification.
	Models also often contain unknown parameters that must be calibrated alongside the missing physics.
	A likelihood based directly on (\ref{eq:system}) would enable joint uncertainty quantification over both the uncertain model structure and these parameters, yielding a fully Bayesian treatment.
	This would also open the door to Bayesian experimental design for missing physics, where the Bayesian framework naturally unifies model discrimination and parameter precision objectives \citep{chaloner, strouwen}.
	
	In summary, we have presented a Bayesian approach to symbolic regression for missing physics that provides principled uncertainty quantification over discovered model structures.
	While the current two-stage approach, first training a neural network, then fitting symbolic expressions, is computationally practical, future work will focus on eliminating this intermediate step to enable fully integrated Bayesian inference and experimental design.
	\bibliography{ifacconf}

\begin{thebibliography}{20}
\providecommand{\natexlab}[1]{#1}
\providecommand{\url}[1]{\texttt{#1}}
\providecommand{\urlprefix}{URL }
\expandafter\ifx\csname urlstyle\endcsname\relax
  \providecommand{\doi}[1]{doi:\discretionary{}{}{}#1}\else
  \providecommand{\doi}{doi:\discretionary{}{}{}\begingroup
  \urlstyle{rm}\Url}\fi

\bibitem[{Chaloner and Verdinelli(1995)}]{chaloner}
Chaloner, K. and Verdinelli, I. (1995).
\newblock Bayesian experimental design: A review.
\newblock \emph{Statistical science}, 273--304.

\bibitem[{Cranmer(2023{\natexlab{a}})}]{cranmer}
Cranmer, M. (2023{\natexlab{a}}).
\newblock Interpretable machine learning for science with pysr and
  symbolicregression. jl.
\newblock \emph{arXiv preprint arXiv:2305.01582}.

\bibitem[{Cranmer(2023{\natexlab{b}})}]{cranmerInterpretableMachineLearning2023}
Cranmer, M. (2023{\natexlab{b}}).
\newblock Interpretable {{Machine Learning}} for {{Science}} with {{PySR}} and
  {{SymbolicRegression}}.jl.
\newblock \doi{10.48550/arXiv.2305.01582}.

\bibitem[{Dandekar et~al.(2020)Dandekar, Chung, Dixit, Tarek, Garcia-Valadez,
  Vemula, and Rackauckas}]{dandekar}
Dandekar, R., Chung, K., Dixit, V., Tarek, M., Garcia-Valadez, A., Vemula,
  K.V., and Rackauckas, C. (2020).
\newblock Bayesian neural ordinary differential equations.
\newblock \emph{arXiv preprint arXiv:2012.07244}.

\bibitem[{Gagnon and Doucet(2020)}]{gagnon}
Gagnon, P. and Doucet, A. (2020).
\newblock Nonreversible jump algorithms for bayesian nested model selection.
\newblock \emph{Journal of Computational and Graphical Statistics}, 30(2),
  312--323.

\bibitem[{Green(1995)}]{green}
Green, P.J. (1995).
\newblock Reversible jump markov chain monte carlo computation and bayesian
  model determination.
\newblock \emph{Biometrika}, 82(4), 711--732.

\bibitem[{Harlim et~al.(2021)Harlim, Jiang, Liang, and Yang}]{harlim}
Harlim, J., Jiang, S.W., Liang, S., and Yang, H. (2021).
\newblock Machine learning for prediction with missing dynamics.
\newblock \emph{Journal of Computational Physics}, 428, 109922.

\bibitem[{Hoffman et~al.(2014)Hoffman, Gelman et~al.}]{hoffman}
Hoffman, M.D., Gelman, A., et~al. (2014).
\newblock The no-u-turn sampler: adaptively setting path lengths in hamiltonian
  monte carlo.
\newblock \emph{J. Mach. Learn. Res.}, 15(1), 1593--1623.

\bibitem[{Jin et~al.(2019)Jin, Fu, Kang, Guo, and Guo}]{jin}
Jin, Y., Fu, W., Kang, J., Guo, J., and Guo, J. (2019).
\newblock Bayesian symbolic regression.
\newblock \emph{arXiv preprint arXiv:1910.08892}.

\bibitem[{Kaiser et~al.(2018)Kaiser, Kutz, and Brunton}]{kaiser}
Kaiser, E., Kutz, J.N., and Brunton, S.L. (2018).
\newblock Sparse identification of nonlinear dynamics for model predictive
  control in the low-data limit.
\newblock \emph{Proceedings of the Royal Society A}, 474(2219), 20180335.

\bibitem[{Keith et~al.(2021)Keith, Khadse, and
  Field}]{keithLearningOrbitalDynamics2021}
Keith, B., Khadse, A., and Field, S.E. (2021).
\newblock Learning orbital dynamics of binary black hole systems from
  gravitational wave measurements.
\newblock \emph{Physical Review Research}, 3(4), 043101.
\newblock \doi{10.1103/PhysRevResearch.3.043101}.

\bibitem[{Koza(1994)}]{koza}
Koza, J.R. (1994).
\newblock Genetic programming as a means for programming computers by natural
  selection.
\newblock \emph{Statistics and computing}, 4, 87--112.

\bibitem[{Philipps et~al.(2024)Philipps, K{\"o}rner, Vanhoefer, Pathirana, and
  Hasenauer}]{philippsNonNegativeUniversalDifferential2024}
Philipps, M., K{\"o}rner, A., Vanhoefer, J., Pathirana, D., and Hasenauer, J.
  (2024).
\newblock Non-{{Negative Universal Differential Equations With Applications}}
  in {{Systems Biology}}.
\newblock \doi{10.48550/ARXIV.2406.14246}.

\bibitem[{Power and Goldman(2019)}]{power}
Power, S. and Goldman, J.V. (2019).
\newblock Accelerated sampling on discrete spaces with non-reversible markov
  processes.
\newblock \emph{arXiv preprint arXiv:1912.04681}.

\bibitem[{Rackauckas et~al.(2020)Rackauckas, Ma, Martensen, Warner, Zubov,
  Supekar, Skinner, Ramadhan, and Edelman}]{rackauckas1}
Rackauckas, C., Ma, Y., Martensen, J., Warner, C., Zubov, K., Supekar, R.,
  Skinner, D., Ramadhan, A., and Edelman, A. (2020).
\newblock Universal differential equations for scientific machine learning.
\newblock \emph{arXiv preprint arXiv:2001.04385}.

\bibitem[{{Rojas-Campos} et~al.(2023){Rojas-Campos}, Stelz, and
  Nieters}]{rojas-camposLearningCOVID19Regional2023}
{Rojas-Campos}, A., Stelz, L., and Nieters, P. (2023).
\newblock Learning {{COVID-19 Regional Transmission Using Universal
  Differential Equations}} in a {{SIR}} model.
\newblock \doi{10.48550/ARXIV.2310.16804}.

\bibitem[{Santana and Costa(2023)}]{santanaEfficientHybridModeling2023}
Santana, V.V. and Costa, E. (2023).
\newblock Efficient hybrid modeling and sorption kinetic model discovery for
  non-linear advection-diffusion-sorption systems: {{A}} systematic scientific
  machine learning approach.

\bibitem[{Strouwen and Miclu{\c{t}}a-C{\^a}mpeanu(2025)}]{strouwen}
Strouwen, A. and Miclu{\c{t}}a-C{\^a}mpeanu, S. (2025).
\newblock Experimental design for missing physics.
\newblock \emph{IFAC-PapersOnLine}, 59(6), 481--486.

\bibitem[{Tierney(1994)}]{tierney}
Tierney, L. (1994).
\newblock Markov chains for exploring posterior distributions.
\newblock \emph{the Annals of Statistics}, 1701--1728.

\bibitem[{Versyck et~al.(1997)Versyck, Claes, and Van~Impe}]{versyck}
Versyck, K.J., Claes, J.E., and Van~Impe, J.F. (1997).
\newblock Practical identification of unstructured growth kinetics by
  application of optimal experimental design.
\newblock \emph{Biotechnology progress}, 13(5), 524--531.

\end{thebibliography}
\end{document}